\documentclass[10pt,twocolumn,letterpaper]{article}

\usepackage[pagenumbers]{cvpr} 

\usepackage[dvipsnames]{xcolor}

\definecolor{cvprblue}{rgb}{0.21,0.49,0.74}
\usepackage[pagebackref,breaklinks,colorlinks,citecolor=cvprblue]{hyperref}
\usepackage{bbm}
\usepackage{pifont}
\def\paperID{*****} 
\def\confName{CVPR}
\def\confYear{2024}

\title{Interactive Continual Learning: Fast and Slow Thinking}

\author{
  Biqing Qi\textsuperscript{1,2,4}, 
  Xinquan Chen\textsuperscript{3},
  Junqi Gao\textsuperscript{3},
  Dong Li\textsuperscript{3},
  Jianxing Liu\textsuperscript{1},
  Ligang Wu\textsuperscript{1,}\thanks{Corresponding authors}, 
  Bowen Zhou\textsuperscript{1,2,4,}\footnotemark[1] \\
  $^1$ Department of Control Science and Engineering, Harbin Institute of Technology, \\
  $^2$ Department of Electronic Engineering, Tsinghua University, \\
  $^3$ School of Mathematics, Harbin Institute of Technology, \\
  $^4$ Frontis.AI, Beijing \\
  {\tt\small \{qibiqing7,xinquanchen0117,gjunqi97,arvinlee826\}@gmail.com,} \\ 
  {\tt\small \{jx.liu,ligangwu\}@hit.edu.cn, \{zhoubowen\}@tsinghua.edu.cn}
  }

\usepackage{multirow}
\usepackage{algorithm}
\usepackage{algorithmic}
\usepackage{caption}
\usepackage{subcaption}
\usepackage{colortbl}

\begin{document}
\maketitle
\begin{abstract}
{Advanced life forms, sustained by the synergistic interaction of neural cognitive mechanisms, continually acquire and transfer knowledge throughout their lifespan. In contrast, contemporary machine learning paradigms exhibit limitations in emulating the facets of continual learning (CL). Nonetheless, the emergence of large language models (LLMs) presents promising avenues for realizing CL via interactions with these models.
Drawing on Complementary Learning System theory, this paper presents a novel Interactive Continual Learning (ICL) framework, enabled by collaborative interactions among models of various sizes. Specifically, we assign the ViT model as System1 and multimodal LLM as System2.
To enable the memory module to deduce tasks from class information and enhance Set2Set retrieval, we propose the Class-Knowledge-Task Multi-Head Attention (CKT-MHA).
Additionally, to improve memory retrieval in System1 through enhanced geometric representation, we introduce the CL-vMF mechanism, based on the von Mises-Fisher (vMF) distribution. 
Meanwhile, we introduce the von Mises-Fisher Outlier Detection and Interaction (vMF-ODI) strategy to identify hard examples, thus enhancing collaboration between System1 and System2 for complex reasoning realization.
Comprehensive evaluation of our proposed ICL demonstrates significant resistance to forgetting and superior performance relative to existing methods. 
Code is available at
\href{https://github.com/Biqing-Qi/Interactive-continual-Learning-Fast-and-Slow-Thinking}{github.com/ICL.}
}
\end{abstract}

\section{Introduction} 
\label{sec:intro}
Advanced life forms exhibit continual learning (CL) and memory formation, facilitated by neural cognitive interactions that enable collaborative knowledge transfer \cite{evans2003two,winocur2007memory,o2002hippocampal}. These underlying mechanisms enhance memory consolidation and utilization, as well as reasoning abilities in advanced life forms \cite{kumaran2016learning,sun2023organizing}. 
However, current machine learning paradigms, particularly neural network-based models, face challenges in achieving CL.
Specifically, neural networks learning from evolving data face a risk known as catastrophic forgetting \cite{cao2023comprehensive}, wherein the integration of new knowledge frequently disrupts existing knowledge, resulting in notable performance degradation \cite{zenke2017continual,lee2020continual,cao2023comprehensive}.

To tackle this challenge, current CL methods strive to preserve and augment knowledge acquired throughout the learning process \cite{kirkpatrick2017overcoming,saha2021gradient,pham2021dualnet,mallya2018piggyback}. 
In CL, rehearsal-based methods \cite{rebuffi2017icarl,saha2021gradient,tang2021layerwise,ECCV22_CL,wang2022learning,wang2022dualprompt}, are the most direct strategy.

However, these methods often ignore the geometric structure \cite{qi2023improving} of memory representations and face challenges in open-class settings. Another perspective includes architecture-based methods \cite{mallya2018piggyback, mallya2018packnet, qin2021bns}, allocating distinct parameters for knowledge encoding from various tasks. Early studies centered on convolution-based architectures \cite{pham2021dualnet}. Recent advancements pivoted towards transformer-based methods like L2P \cite{wang2022learning}, and Dualprompt \cite{wang2022dualprompt}.

From the perspective of Complementary Learning System (CLS) Theory in neurocognitive science \cite{kumaran2016learning}, the current designs of CL frameworks may not be optimal. In a brain-like system, multiple memory modules dynamically maintain a balance between stability and plasticity, with each module possessing predictive capabilities \cite{richards2017persistence,ryan2022forgetting,mermillod2013stability}. However, most advanced CL frameworks lean towards more intuitive systems.
Previous CLS-driven methods \cite{pham2021dualnet, nie2023bilateral} involve the separation and expansion of parameters to facilitate the learning of both fast and slow knowledge. For instance, DualNet \cite{pham2021dualnet} optimizes this process by emphasizing task-specific pattern separation. Similarly, BiMeCo \cite{nie2023bilateral} divides model parameters into two distinct components: a short-term memory module and a long-term memory module. However, these methods \cite{pham2021dualnet, nie2023bilateral} are limited to a single backbone model. In line with CLS principles, this underscores the importance of developing an interactive CL framework between models to consistently achieve higher performance levels.

Meanwhile, recent advancements in large language models (LLMs), as exemplified by ChatGPT \cite{schulman2022chatgpt} and GPT4 \cite{cao2023comprehensive}, have demonstrated remarkable reasoning capabilities. These models can employ chains of thought \cite{wei2022chain} to engage in complex reasoning, much like System2. 
Consequently, it raises an interesting question: Can we integrate intuitive models, such as ViT \cite{dosovitskiy2020image} as System1 alongside LLM-based as System2 to establish an interactive framework for CL?

To response this question, we reevaluated CL by exploring the interaction between ViT (System1) and Multimodal Large Language Model (System2).
In alignment with the current CL setting, our focus is on adapting ViT parameters while keeping System2 parameters stable. System2 is responsible for handling hard examples and facilitates collaboration with System1.
To enable the continual updating of ViT, we introduced the Class-Knowledge-Task Multi-Head Attention (CKT-MHA) module. CKT-MHA utilizes category features and the knowledge of ViT to aid System1 in acquiring task-related knowledge, facilitating knowledge retrieval through Class-Task collections. Furthermore, we introduce the CL-vMF mechanism, which employs von Mises-Fisher distribution modeling to improve memory geometry and enhance retrieval distinguishability through an Expectation-Maximization (EM) update strategy.
This design enables System1 to retain old memory parameters, preventing unnecessary updates and addressing catastrophic forgetting issues.
To realize the coordination between System1 and System2 during their reasoning transitions, we introduce the von Mises-Fisher Outlier Detection and Interaction (vMF-ODI) mechanism for assessing sample difficulty. This mechanism is designed to assist the System1 in adaptive identifying hard examples within each batch. Once identified, these hard examples undergo initial inference by System1, and the resulting prediction outcomes serve as background knowledge for System2 to facilitate more intricate reasoning.

We conduct experiments on various benchmarks, including the demanding Imagenet-R, to validate proposed interactive continual learning (ICL). The results illustrate that ICL significantly mitigates catastrophic forgetting, surpassing state-of-the-art methods. Moreover, it maintains consistently high accuracy across different task. In summary, our contributions are as follows:

\begin{itemize}
    \item We propose an ICL framework  from a novel perspective that emphasizes the interaction between fast intuitive model (ViT) and slow deliberate model (multimodal LLM), aligning with CLS principles.
    \item We propose the CKT-MHA module to acquire task-related information by leveraging category features and small model knowledge. 
   
    \item We propose the CL-VMF mechanism, an optimization strategy guided by VMF distribution modeling with EM strategy updates to enhance the retrieval of geometric memory representations.
  
    \item  
    We propose vMF-ODI, a batch-wise retrieval interaction strategy that enables the adaptive identification of hard examples within each batch, fostering collaborative reasoning between the two systems.
\end{itemize}

\section{Related Works}
\vspace{-3pt}
We discuss three primary categories of CL methods.

\textit{Regularization-based methods} incorporate regularization terms into the loss function to mitigate catastrophic forgetting for previously learned tasks. These methods \cite{kirkpatrick2017overcoming,zenke2017continual,aljundi2018memory,liu2018rotate,lee2020continual,park2019continual}, primarily revolve around the development of metrics for assessing task importance, with additional research efforts dedicated to characterizing the significance of individual features \cite{wang2023comprehensive}. Nevertheless, these methods tend to exhibit reduced performance when applied to more complex datasets.

\textit{Rehearsal based methods}
utilize previous task data to mitigate catastrophic forgetting within limited memory buffers.
Reservoir sampling techniques \cite{chaudhry2019tiny,riemer2018learning}, randomly retain a fixed number of old training samples from each training batch.  Further, \cite{hou2019learning} employs coefficient-based cosine similarity to address sample number imbalances among categories. 
To better recover past knowledge, GEM \cite{saha2021gradient} constructs individual constraints based on old training samples for each task to ensure no increase in their loss. LOGD \cite{tang2021layerwise} decomposes gradients for each task into shared and task-specific components, capitalizing on inter-task information. 
CVT \cite{ECCV22_CL} explores online CL using external attention strategy.

\textit{Architecture-based methods}
 aim at assigning independent parameters for new task data.
 These methods involve strategies such as parameter allocation, model division, and modular network models \cite{mallya2018piggyback, mallya2018packnet,qin2021bns}.
 Previous studies concentrated on specific convolution-based architectures, with DualNet \cite{pham2021dualnet} optimizes memory through separation representations for specific tasks. Recent work focus to transformer-based models. 
 L2P \cite{wang2022learning} enhances integration of  knowledge by treating prompts as optimization parameters. Furthermore, Dualprompt \cite{wang2022dualprompt} enhances knowledge memory by constructing dual orthogonal prompt spaces.

\section{Methodology}
\begin{figure*}[h]
\vspace{-15pt}
  \centering
  \includegraphics[width=0.68\textwidth]{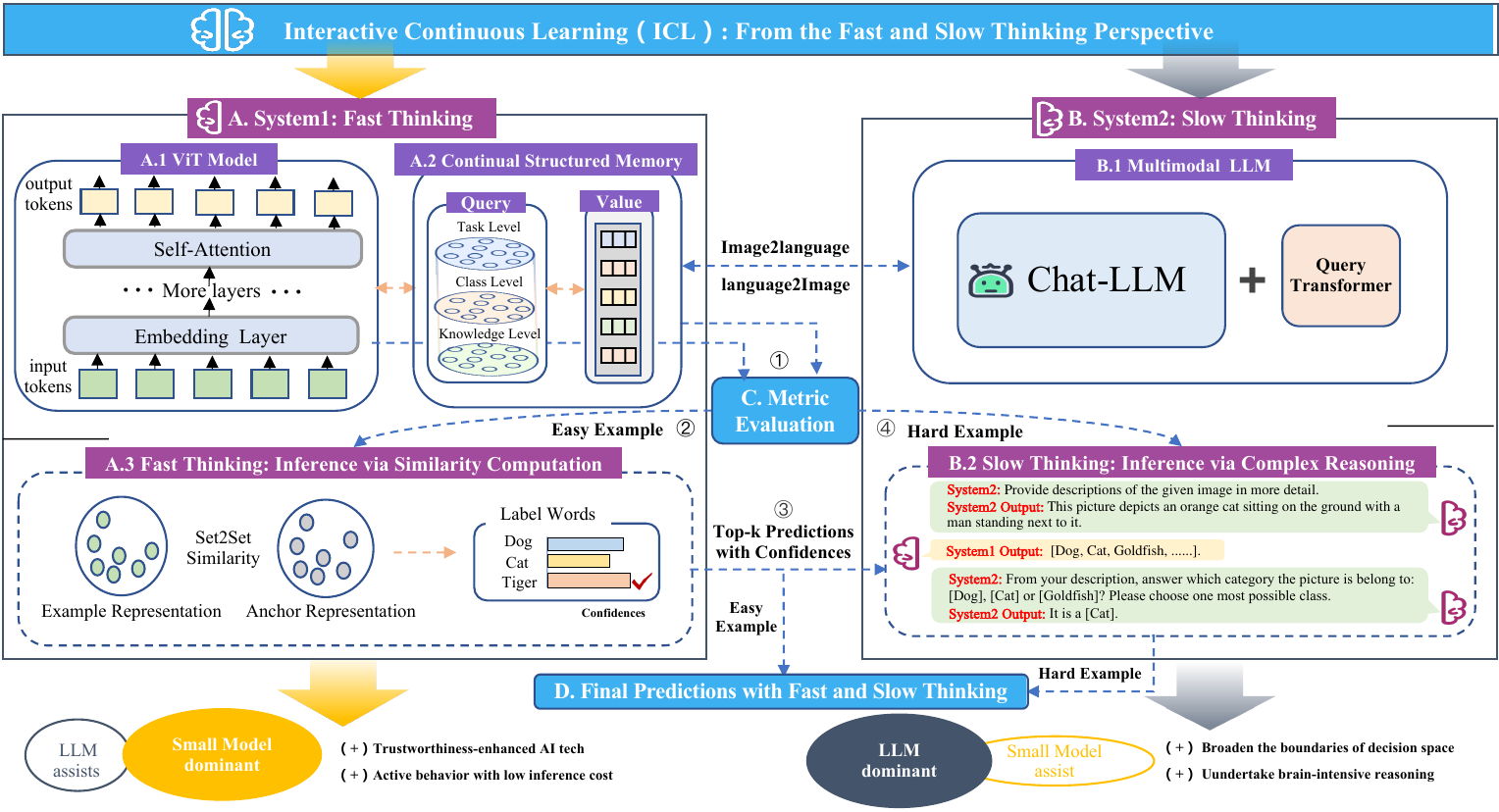} 
 \caption{Comprehensive Training and Testing Illustration. In the training phase: we propose the CKT-MHA unified storage module for System1. And then use memory selection and updates through our CL-VMF mechanism with the EM strategy to optimize CL for the small model ViT. In the inference phase: 1) The process begins by assessing sample complexity using proposed vMF-ODI in System 1. 2) The System1 then swiftly generates inferential predictions.
3) If test samples surpass a complexity threshold, we activate collaborative inference. Specifically, the predictive results from System1 is used as background knowledge to narrow the scope of inference. 
4) Subsequently, complex reasoning through the multimodal LLM is applied to achieve the final prediction.}
  \label{fig:multi_agent}
  \vspace{-8pt}
\end{figure*}
\subsection{Problem Setup} 
A standard paradigm for CL can be defined by a set of task descriptors $t\in\mathcal T$, and the corresponding distribution $p_t(\boldsymbol x,y)$ for each task. The task-specific dataset $\mathcal D_{t}:=\{(\boldsymbol x_i^t, y_i^t)\mid (\boldsymbol x_i^t, y_i^t)\sim p_t, i\in[N_t]\}$, where $N_t$ represents the number of samples in the training set of the $t$-th task. The dataset is drawn i.i.d. from the sample space $\mathcal X^t\times\mathcal Y^t\in\mathcal X\times \mathcal Y$. During the training phase, the training samples are sequentially fed as inputs, following the task descriptors from $0$ to $|\mathcal T|$. 
In formulating the ICL framework, we provide definitions for System1 and System2, respectively.
System1 is instantiated by the model $f_{\boldsymbol \theta}(\cdot):\mathcal X\mapsto \mathbb R^d$ parameterized by $\boldsymbol \theta$ which updates its parameters to $\boldsymbol \theta_t$ in the $t$-th task. Our objective is to determine $\boldsymbol \theta_{|\mathcal T|}=\arg\min_{\boldsymbol \theta_{|\mathcal T|}}\mathbb E_{t\in\mathcal T}\mathbb E_{(\boldsymbol x^t,y^t)}\left [\ell(f_{\boldsymbol \theta_{|\mathcal T|}}(\boldsymbol x^t), y_i^t)\right ]$, ensuring the memory capacity of the System1. Here $\ell(\cdot, y^t)$ represents loss function, with $y^t\in\mathcal Y^t$. The system1 utilizes a memory buffer $\mathcal M$ and updates $\boldsymbol \theta_{t-1}$ to $\boldsymbol \theta_{t}$ by using $\mathcal D_{t}\cap\mathcal M$. Furthermore, System2 is instantiated with a Multimodal LLM represented as $g_{\boldsymbol{\psi}}$ to model complex reasoning abilities.
To enable collaborative inference, System2 must handle hard samples $\tilde{\boldsymbol{X}}$ that System1 struggles with, maximizing the probability $\mathbb P_{\tilde{\boldsymbol{x}}\in\tilde{\boldsymbol{X}}}\left(\tilde y\ne\arg\max_{i}f_{\boldsymbol{\theta}}(\tilde{\boldsymbol{x}})_{i}\right)$. This requires the ability of System1 to filter out these samples, thereby improving the inference of System2 by leveraging predicted results from System1
$I_{f_{\boldsymbol\theta}}$ for second-stage inference:
\vspace{-3pt}
\begin{equation}
\vspace{-3pt}
    \hat{\tilde y} = \mathcal S^{-1}\left(g_{\boldsymbol\psi}\left(\tilde{\boldsymbol x};\mathcal S\left(I_{f_{\boldsymbol\theta}}(\tilde{\boldsymbol x})\right)\right)\right),
\end{equation}
here $\mathcal S$ represents label to prompt operation, while $\mathcal S^{-1}$ represents its inverse operation. System2 is expected to produce an output $S(\tilde y)$ that minimizes $ -\log p_{\boldsymbol\psi}\left(\mathcal S(\tilde y)\mid\tilde{\boldsymbol x};\mathcal S\left(I_{f_{\boldsymbol \theta}}(\tilde{\boldsymbol x})\right)\right)$. Next, we will present detailed designs for each component of the ICL framework and discuss optimization strategies.
\subsection{Query and Value Memory for System1}

In general, deep neural networks implicitly encode data memories within their parameters, and unnecessary parameter changes can result in memory degradation.
The prevailing methods to deploying models in downstream tasks involves utilizing pre-trained feature extractors and introducing new parameters for adaptation, which has been demonstrated to be effective in CL setting \cite{wang2022learning, wang2022dualprompt}. Nevertheless, these methods face challenge when the precise number of classes in the downstream task is uncertain. Moreover, updating all parameters of the classification head for each new task worsens the issue of forgetting. To address these challenges, we propose separating the model's parameters into two distinct groups: value memory parameters, denoted as $\mathcal Z={\mathcal Z_1, \mathcal Z_{2}, ...}$ with class-specific representations $\boldsymbol z^{y^t}\in\mathcal Z_t$, and query memory parameters represented by $\boldsymbol \theta$. This decoupling strategy enhances operational flexibility. In simpler terms, we envision $\mathcal Z$ as a collection of class-specific value memory variables, ensuring that
\begin{itemize}
    \item Value memory parameters $\mathcal Z$ can be augmented as the number of tasks increases, i.e., there is a memory increment. When training on task $t$, $\mathcal Z$ will be updated from $\{\mathcal Z_1, ..., \mathcal Z_{t-1}\}$ to $\{\mathcal Z_1, ..., \mathcal Z_{t}\}$.
    \item The relevant value memory for each class is only updated when necessary, i.e., the value memory $z^{y^t}$ will not be updated if the input does not contain class $y^t$, thus $\boldsymbol{z}^{y^{t-1}}$ after trained on task $t$ equals to that trained before training on task $t$.
\end{itemize}
Then we can ensure that the System1 can be equipped with persistent value memory of old data and update query parameters using a portion of the memory buffer as rehearsal samples, thus can effectively handle old data. At the same time, this enables System1 to be unconstrained by a predefined number of classes and allows for more flexible memory allocation for new tasks.
\vspace{-10pt}
\paragraph{Interactive Query and Value Memory with CKT-MHA
 } Firstly, we design such a memory module for System1. Specifically, we propose a Set2Set memory retrieval mechanism to further enhance memory retrieval stability. This mechanism involves first obtaining class (cls) information via a projector $f_{\boldsymbol\theta_c}$ parameterized by $\boldsymbol\theta_c$: $\boldsymbol \xi_{c}=f_{\boldsymbol\theta_c,\boldsymbol\varphi}(\boldsymbol x)\in\mathbb R^{L\times d_c}$, $L$ is the obtained token length. Here we use the pre-trained ViT as image feature extractor $f_{\boldsymbol \varphi}$, followed by an query interactor $f_{\boldsymbol \theta}$, parameterized by $\boldsymbol \theta$, introduced for memory matching. This results in $f_{\boldsymbol \theta, \boldsymbol \varphi}=f_{\boldsymbol \theta}\circ f_{\boldsymbol\varphi}$. Subsequently, we utilize $\boldsymbol \xi_{c}$ and pretrained knowledge $\boldsymbol \xi_k = f_{\boldsymbol\varphi}(\boldsymbol x)\in\mathbb R^{L\times d_c}$ to construct the Class-Knowledge-Task Multi-Head Attention (CKT-MHA) to capture the task information corresponding to the class:
 \vspace{-5pt}
\begin{align}
 \vspace{-5pt}
&\boldsymbol h_{\tau,i}  = \text{Attention}_{\boldsymbol \theta_{SA}}\left( \boldsymbol \xi_{c}^{[:, R_i]}, \boldsymbol \xi_{k}^{[:,R_i]}, f_{\boldsymbol\theta_\tau}(\boldsymbol \xi_k)^{[:,R_i]} \right),\\
&\boldsymbol \xi_\tau = f_{\boldsymbol\theta_o}\left(\text{Concat}\left[\boldsymbol h_{\tau,1},\dots, \boldsymbol h_{\tau,N_h}\right]\right)\in\mathbb R^{L\times\tau},
\end{align}
where $N_h$ is the head number of MHA, $R_i=[(i-1)*\frac{d_c}{N_h}, i*\frac{d_c}{N_h}]$ is the corresponding attention head interval. Then we combine the class and task features set to obtain the classification feature: $\boldsymbol\xi = \text{Concat }[\frac{1}{L}\sum_{l=1}^L\boldsymbol\xi_\tau^l, \frac{1}{L}\sum_{l=1}^L\boldsymbol\xi_c^l]$, where $\boldsymbol\xi_\tau^l$ denotes the $l$-th token of $\boldsymbol\xi_\tau$. Consequently, the interactor parameters are denoted as $\boldsymbol\theta =\{\boldsymbol\theta_c, \boldsymbol\theta_\tau, \boldsymbol\theta_{SA},\boldsymbol\theta_o\}$. Finally, we set a group of task-specified value memory vector $\boldsymbol z_\tau\in\mathcal Z^\tau$ and class-specified $\boldsymbol z_c\in\mathcal Z^c$ corresponding to the class to form the final value memory variable $\boldsymbol z$ for retrieval: $\boldsymbol{z} = \text{Concat }[\boldsymbol z_\tau, \boldsymbol z_c]$. During the inference stage, the proposed task-class Set2Set retrieval is performed from value memory as: $y_{\text{pred}}=\arg\max_{\boldsymbol z\in\mathcal Z}p_{\boldsymbol{\theta},\boldsymbol \varphi}(\boldsymbol z|\boldsymbol x)$.

With decoupled parameters, we also need to decouple the updates of query memory parameters and value memory parameters to ensure the flexibility of adjusting these two parts of parameters, avoiding unnecessary updates to value memory parameters, which prevents biases in memory variables for future tasks and mitigates catastrophic forgetting. Furthermore, 
prioritizing value memory parameter optimization guides the optimization of query parameters. Is it possible to achieve such decoupled optimization while ensuring a consistent optimization objective? 
The inherent nature of EM algorithm provides a framework for meeting these optimization requirements. Specifically, considering the Maximum Likelihood Estimation (MLE) under a classification setting, with the probability of task $t$ denoted as $\mathbb P(t)$, the objective can be expressed as follows:
\vspace{-3pt}
\begin{equation}
\vspace{-5pt}
\underset{\boldsymbol\theta_{|\mathcal T|}}{\text{Minimize}}\sum_{t}\mathbb P(t)\mathbb E_{\boldsymbol x^t}\left [-\log p_{\boldsymbol{\theta_{|\mathcal{T}|}}}(y^t|\boldsymbol x^t)\right ].
\end{equation}
For the sake of simplicity in the framework, here we only consider the parameters that are being updated. During training on task $t$, the System1 aims to determine $\boldsymbol\theta_{t}=\arg\min_{\boldsymbol\theta_{t}}\mathbb E_{\boldsymbol x^t}\left [-\log p_{\boldsymbol{\theta_{t}}}(y^t|\boldsymbol x^t)\right ]$. To achieve this, we model value memory $\boldsymbol z\in\mathcal Z$ as hidden variables, alongside the hidden distribution $q(\boldsymbol z|\boldsymbol x)$ which satisfies $\sum_{\boldsymbol z\in\mathcal Z}q(\boldsymbol z|\boldsymbol x)=1$. This allows us to obtain
\begin{align}
\vspace{-5pt}
\resizebox{0.42\textwidth}{!}{$\log p_{\boldsymbol{\theta_{t}}}(y^t|\boldsymbol x^t) = \log \left [\frac{p_{\boldsymbol{\theta_{t}}}(y^t,\boldsymbol z| \boldsymbol x^t)}{q_(\boldsymbol z|\boldsymbol x^t)}\right ] - \log \left [\frac{p_{\boldsymbol{\theta_{t}}}(\boldsymbol z|y^t, \boldsymbol x^t)}{{q_(\boldsymbol z|\boldsymbol x^t)}}\right ]$},
\vspace{-5pt}
\end{align}
taking the expectation with respect to $q(\boldsymbol z|\boldsymbol x)$ on both sides,
\vspace{-5pt}
\begin{align}
\log p_{\boldsymbol{\theta_{t}}}(y^t|\boldsymbol x^t)& = \sum_{\boldsymbol z\in \mathcal Z}q(\boldsymbol z|\boldsymbol x^t)\log \left [\frac{p_{\boldsymbol{\theta_{t}}}(y^t,\boldsymbol z| \boldsymbol x^t)}{q(\boldsymbol z|\boldsymbol x^t)}\right ],\\
&+ \sum_{\boldsymbol z\in \mathcal Z}q(\boldsymbol z|\boldsymbol x^t)\log \left [\frac{{q(\boldsymbol z|\boldsymbol x^t)}}{p_{\boldsymbol{\theta_{t}}}(\boldsymbol z|y^t, \boldsymbol x^t)}\right ],
\vspace{-5pt}
\end{align}
the first term of R.H.S is referred to as the Evidence Lower Bound $\mathcal L_{\text{EL}}\left(p_{\boldsymbol{\theta_{t}}}(y^t,\boldsymbol z| \boldsymbol x^t), q(\boldsymbol z|\boldsymbol x^t)\right)$, the second term is $\mathbf{KL}\left(q(\boldsymbol z|\boldsymbol x^t)\|p_{\boldsymbol{\theta_{t}}}(\boldsymbol z|y^t, \boldsymbol x^t)\right )$. Since the KL divergence is non-negative, we can achieve maximum likelihood by iteratively updating $\mathcal Z$ and $\boldsymbol \theta_t$ using the Generalized Expectation-Maximization algorithm (GEM) as follows:
\vspace{-5pt}
\begin{equation}
    \resizebox{0.43\textwidth}{!}{$\mathcal Z^{(i+1)}=\arg\min_{\mathcal Z}\mathbb E_{\boldsymbol x^t}\left[\mathbf{KL}\left(q(\boldsymbol z^{(i)}|\boldsymbol x^t)\|p_{\boldsymbol{\theta_{t}^{(i)}}}(\boldsymbol z^{(i)}|y^t, \boldsymbol x^t)\right )\right]$},
\vspace{-2pt}
\end{equation}
\begin{equation}
    \resizebox{0.43\textwidth}{!}{$\boldsymbol{\theta}^{(i+1)}_{t}=\arg\max_{\boldsymbol\theta} \mathbb E_{\boldsymbol x^t}\left[\mathcal L_{\text{EL}}\left(p_{\boldsymbol{\theta^{(i)}_{t}}}(y^t,\boldsymbol z^{(i+1)}| \boldsymbol x^t), q(\boldsymbol z^{(i+1)}|\boldsymbol x^t)\right)\right]$}.
\end{equation}
In the context of supervised learning, we can define $\boldsymbol z$ as class-specific $\boldsymbol z^{y^t}\in\mathcal Z_t$. The prior form of $q$ is $q(\boldsymbol z|\boldsymbol x^t) = \sum_{\boldsymbol z\in\mathcal Z}\mathbbm 1(\boldsymbol z = \boldsymbol z^{y^t})$. Note that in this case, 
\vspace{-4pt}
\begin{equation}
p_{\boldsymbol{\theta}_{t}}(\boldsymbol z|y^t, \boldsymbol x^t)=p_{\boldsymbol{\theta}_{t}}(\boldsymbol z^{y^t}| \boldsymbol x^t)=p_{\boldsymbol{\theta}_{t}}(\boldsymbol z,y^t| \boldsymbol x^t).
\vspace{-4pt}
\end{equation}
Hence, we can express the two distinct objectives in eq.(8) and (9) as a unified objective:
\vspace{-4pt}
\begin{equation}
\boldsymbol\theta_t^*,\mathcal Z^*_t=\arg\min_{\boldsymbol\theta,\mathcal Z}\mathbb E_{\boldsymbol x^t}\left[-\log p_{\boldsymbol{\theta}_{t}}(\boldsymbol z^{y^t}|\boldsymbol x^t)\right].
\vspace{-4pt}
\end{equation}
When training on a sequence of tasks, if we directly optimize as in eq.(4) for the current task's optimal $\boldsymbol \theta_t^*=\arg\min_{\boldsymbol \theta}\mathbb E_{\boldsymbol x^t}\left [-\log p_{\boldsymbol{\theta}_{t}}(y^t|\boldsymbol x^t)\right ]$, as is typical, it will inevitably lead to $\boldsymbol \theta^*_{t}\ne\boldsymbol \theta^*_{t-1}$, resulting in catastrophic forgetting. However, our decoupled optimization strategy ensures that the value memory parameters remain optimal for their corresponding tasks, and with the integration of rehearsal, it guarantees that the optimization of query memory parameters can be consistently guided by such value memory parameters of old tasks, thereby mitigating catastrophic forgetting while continually adapting to new data.
Next, we describe how to model $p_{\boldsymbol{\theta}_{t}}(\boldsymbol z^{y^t}|\boldsymbol x^t)$ in eq.(11). 

\subsection{Optimizing Memory via CL-vMF} 
\textbf{Modelling Posterior with vMF Distribution.} To assure value memory vectors more discriminative that facilitate more explicit memory retrieval, necessitating the construction of more separable geometric relationships for them. Introducing improved geometric relationships for value memory vectors further ensures that the query features of each class are more separable and closer to the class center when combined with the EM updating strategy. This also aids in more effectively filtering out outliers, laying the groundwork for screening hard samples for System1, as discussed in Section 3.4. Therefore, we opt for the von Mises-Fisher (vMF) distribution, which naturally excels in modeling geometric relationships in high-dimensional spaces \cite{10.5555/1046920.1088718, levy-etal-2015-improving, Qi2023ImprovingRO} and has demonstrated effectiveness in downstream tasks \cite{TrioZhang, MengSZH19}. Its probability density function is as follows:
\vspace{-5pt}
\begin{equation}
    p_{\text{vMF}}(\boldsymbol x;\boldsymbol \mu,\kappa)=C_d(\kappa)\exp(\kappa\left \langle \boldsymbol \mu,\boldsymbol x \right \rangle ),
\vspace{-5pt}
\end{equation}
where the correlated sample space is defined as $\{\boldsymbol x|\boldsymbol x\in \mathbb{R}^d,\|\boldsymbol x\|=1\}$. $\boldsymbol \mu \in \mathbb R^d$ represent the mean direction, while $\kappa$ is the concentration parameter. The constant $C_d(\kappa)$ is only related to $\kappa$ and $d$.
By setting normalized value memory 
parameter $\boldsymbol{z}^{y^t}/\|\boldsymbol{z}^{y^t}\|$ to the mean direction, we model the distribution of the normalized feature $f_{\boldsymbol \theta, \boldsymbol\varphi}(\boldsymbol x^t)/\|f_{\boldsymbol \theta, \boldsymbol\varphi}(\boldsymbol x^t)\|$ for class $y^t$ as a vMF distribution with probability density $C_d(\kappa)\exp(\kappa\left \langle  \frac{\boldsymbol z^{y^t}}{\|\boldsymbol z^{y^t}\|},\frac{f_{\boldsymbol \theta,\boldsymbol\varphi}(\boldsymbol x^t)}{\|f_{\boldsymbol \theta, \boldsymbol\varphi}(\boldsymbol x^t)\|} \right \rangle )$. By utilizing the constructed probability density for Bayesian discrimination, we can ascertain the form of the posterior probability $p_{\boldsymbol{\theta}_{t}, \boldsymbol\varphi}(\boldsymbol z^{y^t}|\boldsymbol x^t)$ as:
\vspace{-5pt}
\begin{equation}
p_{\boldsymbol{\theta}_{t}, \boldsymbol\varphi}(\boldsymbol z^{y^t}|\boldsymbol x^t)=\frac{\exp(\kappa\left \langle  \frac{\boldsymbol z^{y^t}}{\|\boldsymbol z^{y^t}\|},\frac{f_{\boldsymbol \theta, \boldsymbol\varphi}(\boldsymbol x^t)}{\|f_{\boldsymbol \theta, \boldsymbol\varphi}(\boldsymbol x^t)\|} \right \rangle )}{\sum_{\boldsymbol z^\prime\in\mathcal Z}\exp(\kappa\left \langle  \frac{\boldsymbol z^\prime}{\|\boldsymbol z^\prime\|},\frac{f_{\boldsymbol \theta, \boldsymbol\varphi}(\boldsymbol x^t)}{\|f_{\boldsymbol \theta, \boldsymbol\varphi}(\boldsymbol x^t)\|} \right \rangle ).}
\vspace{-2pt}
\end{equation}
When a sample pair of a new class $(x^{t+1}, y^{t+1})\in\mathcal D^{t+1}$ is inputted, a new $\boldsymbol z^{y^{t+1}}\notin \mathcal Z$ can be assigned to it for memory expansion, thereby avoiding the limitation imposed by the predefined number of classes, i.e., class-free. Combined with the proposed memory module, we introduce CL-vMF, a class-free memory retrieval mechanism that accommodates new classes when the total number is unknown, it retrieves from memories based on the vMF posterior and ensures that class-specific value memory parameters are only updated when needed. 
\vspace{-10pt}
\paragraph{Implementation of CL-vMF.}
We set the value memory vectors $\mathcal Z$ as learnable memory embeddings, enabling the model to use them for retrieval using the vMF posterior to ascertain a class of sample. The process resembles constructing a new "hippocampus" within the pre-trained model by utilizing $\mathcal Z$ and learning query memory interactions via $\boldsymbol \theta$ to adapt and employ the value memory. In the training phase, upon encountering each new class $y$, we allocate a $\boldsymbol z^y$, incorporating it into $\mathcal Z$. To ensure updating the value memory parameter $\boldsymbol z^y$ exclusively when the input contains a sample of class $y$, we introduce the subsequent batchwise vMF loss to maximize likelihood:
\vspace{-3pt}
\begin{equation}
    \mathcal L_{\text{vMF}}(\mathcal B_{i}^t)=\frac{1}{|\mathcal B_{i}^t|}\sum_{(\boldsymbol x^t,y^t)\in\mathcal B_{i}^t}-\log\left[p^{\mathcal B^t_i}_{\boldsymbol{\theta},\boldsymbol \varphi}(\boldsymbol z^{y^t}|\boldsymbol x^t)\right],
\vspace{-4pt}
\end{equation}
where $\mathcal B^t_i\in\mathcal D^t$ is the $i$-th input batch of task $t$ and
\vspace{-3pt}
\begin{equation}
    \resizebox{0.41\textwidth}{!}{$p^{\mathcal B^t_i}_{\boldsymbol{\theta},\boldsymbol \varphi}(\boldsymbol z^{y^t}|\boldsymbol x^t)=
\frac{\exp(\kappa\left \langle \Pi_\mathcal N(\boldsymbol z^{y^t}),\Pi_\mathcal N\left(\boldsymbol\xi^t\right)\right \rangle )}{\sum_{y\in\mathcal Y_i^t}\exp(\kappa\left \langle  \Pi_\mathcal N\left(\boldsymbol z^y\right),\Pi_\mathcal N\left(\boldsymbol\xi^t\right) \right \rangle )}$},
\vspace{-3pt}
\end{equation}
where $\boldsymbol\xi^t=f_{\boldsymbol \theta, \boldsymbol\varphi}(\boldsymbol x^t)$, $\mathcal Y^t_i=\text{Unique}(\{y\}_{(\boldsymbol x, y)\in \mathcal B^t_i})$ is the set of classes that appears in $\mathcal B^t_i$, $\Pi_\mathcal N(\cdot):=\frac{(\cdot)}{\|\cdot\|}$ denotes the projection operator. $\kappa$ is set as a hyperparameter.
Hence, our training approach involves alternating between the EM steps based on eq.(8) and eq.(9)  throughout the training process. Nonetheless, when delving deeper into the optimization process, it becomes imperative to consider the gradient of $\Pi_\mathcal N(\boldsymbol z^{y^t})$:
\vspace{-3pt}
\begin{align}
&\nabla_{\Pi_\mathcal N(\boldsymbol z^{y})}\left[-\log\left(p^{\mathcal B^t_i}_{\boldsymbol{\theta},\boldsymbol \varphi}(\boldsymbol z^{y^t}|\boldsymbol x^t)\right)\right ]\\
&=\left\{\begin{matrix}
  & \left[p^{\mathcal B^t_i}_{\boldsymbol{\theta},\boldsymbol \varphi}(\boldsymbol z^{y^t}|\boldsymbol x^t)-1\right ]\kappa \Pi_\mathcal N\left(\boldsymbol\xi^t\right),\quad y=y^t\\
  & p^{\mathcal B^t_i}_{\boldsymbol{\theta},\boldsymbol \varphi}(\boldsymbol z^{y}|\boldsymbol x^t)\kappa \Pi_\mathcal N\left(\boldsymbol\xi^t\right),\quad y\ne y^t
\end{matrix}\right.\nonumber,
\vspace{-3pt}
\end{align}
thus when $p^{\mathcal B^t_i}_{\boldsymbol{\theta},\boldsymbol \varphi}(\boldsymbol z^{y^t}|\boldsymbol x^t)\rightarrow 1$, the gradient for $\Pi_\mathcal N(\boldsymbol z^{y})$ ($y = y^t$) will tend to zero, meanwhile, as $p$ approaches $0$, $\nabla_{\Pi_\mathcal N(\boldsymbol z^{y})}\left[-\log\left(p^{\mathcal B^t_i}_{\boldsymbol{\theta},\boldsymbol \varphi}(\boldsymbol z^{y^t}|\boldsymbol x^t)\right)\right ](y\ne y^t)$ will also tend to zero. Likewise, 
for $\Pi_\mathcal N\left(\boldsymbol\xi^t\right)$ we have:
\vspace{-3pt}
\begin{align}
&\nabla_{\Pi_\mathcal N\left(\boldsymbol \xi^t\right)}\left[-\log\left(p^{\mathcal B^t_i}_{\boldsymbol{\theta},\boldsymbol \varphi}(\boldsymbol z^{y^t}|\boldsymbol x^t)\right)\right ]\\
&=\kappa\left[\sum_{y\in\mathcal Y^{t}_i} p^{\mathcal B^t_i}_{\boldsymbol{\theta},\boldsymbol \varphi}(\boldsymbol z^{y}|\boldsymbol x^t)\Pi_\mathcal N(\boldsymbol z^{y})-\Pi_\mathcal N(\boldsymbol z^{y^t})\right]\nonumber
\vspace{-6pt}.
\end{align}
Note that when $p^{\mathcal B^t_i}_{\boldsymbol{\theta},\boldsymbol \varphi}(\boldsymbol z^{y^t}|\boldsymbol x^t)\rightarrow 1$, 
\vspace{-3pt}
\begin{equation}
\sum_{y\in\mathcal Y^{t}_i} p^{\mathcal B^t_i}_{\boldsymbol{\theta},\boldsymbol \varphi}(\boldsymbol z^{y}|\boldsymbol x^t)\Pi_\mathcal N(\boldsymbol z^{y}){\rightarrow} \Pi_\mathcal N(\boldsymbol z^{y^t}).
\vspace{-5pt}
\end{equation}
This means that $\nabla_{\Pi_\mathcal N\left(\boldsymbol \xi^t\right)}\left[-\log\left(p^{\mathcal B^t_i}_{\boldsymbol{\theta},\boldsymbol \varphi}(\boldsymbol z^{y^t}|\boldsymbol x^t)\right)\right ]$ will tend to zero.
To ensure stable gradients during training and achieve consistent loss reduction, we introduce a gradient stabilization loss
\vspace{-3pt}
\begin{equation}\resizebox{0.41\textwidth}{!}{$
    \ell^\delta_{\text{GS}}(\boldsymbol x^t, y^t)=\ell_{\text{margin}}^\delta\left(\boldsymbol z^{y^t},\boldsymbol\xi^t\right)+\sum_{y\in\mathcal Y^t_i,y\ne y^t} \ell_{\text{margin}}^\delta\left(-\boldsymbol z^{y},\boldsymbol\xi^t\right)$},
\vspace{-3pt}
\end{equation}
where
\vspace{-3pt}
\begin{equation}\resizebox{0.41\textwidth}{!}{$
    \ell_{\text{margin}}^\delta\left(\boldsymbol z^y,\boldsymbol\xi^t\right)=\max(\left|1-\left \langle \Pi_\mathcal N\left(\boldsymbol z^y\right),\Pi_\mathcal N\left(\boldsymbol\xi^t\right) \right \rangle\right|-\delta)$},
\vspace{-3pt}
\end{equation}
which provides a constant gradient as compensation, with $\delta$ regulating the threshold of gradient compensation. When the loss becomes small, no further compensation is provided due to the instability of the zero point for the absolute value function. Similarly, we incorporate the batchwise gradient stabilization loss $\mathcal L_{\text{GS}}^\delta(\mathcal B_{i}^t) = \frac{1}{|\mathcal B_{i}^t|}\sum_{(\boldsymbol x^t,y^t)\in\mathcal B_{i}^t}\ell_{\text{GS}}(\boldsymbol x^t, y^t)$ into the objective, leading to the overall loss:
\vspace{-2pt}
\begin{equation}
    \mathcal L\left(\mathcal B^t_i\right)=\mathcal L_{\text{vMF}} \left(\mathcal B^t_i\right)+ \lambda \mathcal L^\delta_{\text{GS}}\left(\mathcal B^t_i\right),
\vspace{-2pt}
\end{equation}
here, $\delta$ and $\lambda$ are hyperparameters.

The CL-vMF model possesses several advantageous features, including incremental value memory and the ability to handle an arbitrary number of classes. As a result, there is no need to retrain the classification head even when the number of classes exceeds the pre-defined limit. Since the value memory parameters $\mathcal Z^t$ are frozen after the completion of task $t$,  i.e. all memories $\mathcal Z^1, \dots, \mathcal Z^{t-1}$ remain unchanged both before and after training task $t$. This guarantees stable and persistent value memory. Moreover, the storage cost of class value memory parameters is calculated as $d_{\boldsymbol z_\tau}\times N_C + d_{\boldsymbol z_c}\times |\mathcal T|$, where $d_{\boldsymbol z_c}, d_{\boldsymbol z_\tau}$ and $N_C$ represent the dimensions of $\boldsymbol{ z_\tau}, \boldsymbol{z}_c$ and the number of classes, respectively. Importantly, this implies that the storage cost of value memory parameters scales linearly with $N_C$.

\subsection{Collaborative Inference: System1 and System2}
\vspace{-5pt}
To align with the CLS, we aim for the ICL framework to activate System2 when System1 fails to perform fast thinking, i.e. when encounters hard samples. This activation leverages the complex reasoning capabilities of System2 to achieve collaborative inference. Specifically, we use MLLM to instantiate $g_{\boldsymbol \psi}$.
And we propose a hard sample filtering strategy, vMF-ODI, to screen data that challenges the System1. Specifically, we use batchwise normalization to filter outliers, thus identifying hard sample set $\tilde {\boldsymbol{X}}$:
 \vspace{-5pt}
\begin{equation}
    \tilde {\boldsymbol{X}_i} = \left\{(\tilde{\boldsymbol x},\tilde y )\in\mathcal B_i\mid (\nu-\bar\nu_{\mathcal B_i})/\sigma_{\mathcal B_i}<\alpha\right\},
    \vspace{-4pt}
\end{equation}
where $\nu = \frac{\left \langle f_{\boldsymbol \theta,\boldsymbol\varphi}(\boldsymbol x), \boldsymbol z^{\hat y} \right \rangle }{\|f_{\boldsymbol \theta,\boldsymbol\varphi}(\boldsymbol x)\|\|\boldsymbol z^{\hat y}\|}$, $\hat y$ is the predicted label, $\nu_{\mathcal B_i}=\frac{1}{|\mathcal B_i|}\sum_{j\in \mathcal B_i}\nu_j$, $\sigma=\sqrt{\frac{1}{|\mathcal B_i|-1}\sum_{j\in{\mathcal B_i}}(\nu_j-\bar\nu)^2}$ and 
$\alpha$ is a detection threshold. For the filtered hard samples, we utilize the TopK outputs $[f_{\boldsymbol \theta,\boldsymbol\varphi}(\tilde{\boldsymbol x})]^K$ from the System1 to construct an inquiry-based language prompt $\mathcal S\left([f_{\boldsymbol \theta,\boldsymbol\varphi}(\tilde{\boldsymbol x})]^K\right)$. The operation $\mathcal S$ converts labels to language and prompts the System2 to perform reasoning and rank the results based on the given context, and finally gives the prediction $\hat{\tilde{y}}$ like we stated in eq. (1). After completing the inference in stage 2, if the System2 provides a precise answer, we will use that answer. Otherwise, we will rely on the judgment result from the System1. This interactive scheme suggests the possibility of using fine-tuning strategies, such as LoRA \cite{hu2021lora}, to minimize $-\log p_{\boldsymbol\psi}\left(\mathcal S(y)\mid\boldsymbol x;\mathcal S\left([f_{\boldsymbol \theta,\boldsymbol\varphi}(\boldsymbol x)]^K\right)\right)$ with respect to $\psi$. 
This adjustment would help align the System2 with the System1. Consistent with existing CL setups, we focus on the parameter updates of System1.
Detailed algorithmic description can be found in Appendix A. During the training process, we perform $T$ iterations of EM alternating steps for each task, with updates only applied to $\boldsymbol{\theta}$ in subsequent updates for that task.

\vspace{-5pt}
\section{Experiments}
\vspace{-3pt}
\begin{figure*}[t]
  \vspace{-7pt}
  \centering
  \includegraphics[width=0.82\linewidth]{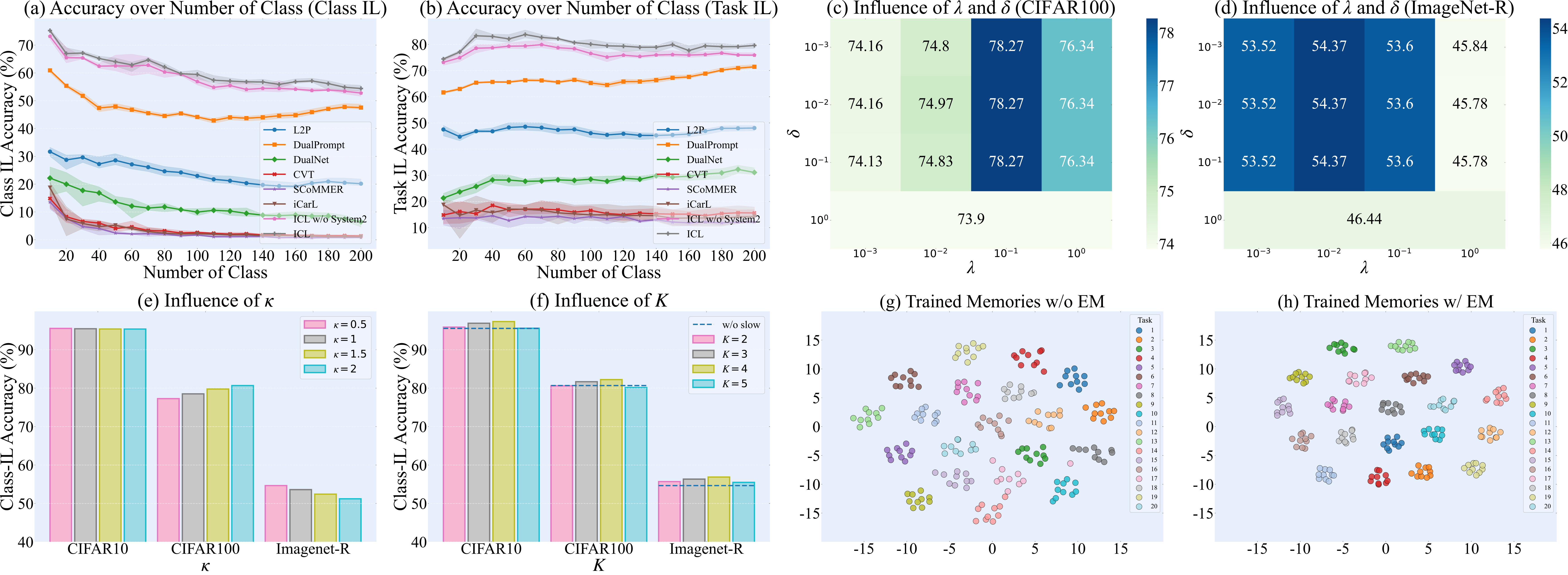}
 \caption{Further analysis of the proposed ICL. (a) and (b) are forgetting curves of different methods on the ImageNet-R in the Class IL and Task IL scenario respectively. (c) and (d) are the impact of regularization parameters $\lambda$ and $\delta$ on CIFAR100 and ImageNet-R respectively. (e) The impact of concentration $\kappa$, is evaluated at values of 0.5, 1, 1.5, and 2, respectively. (f) The impact of the number of category choices $K$ in the prompt, is evaluated at values of 2, 3, 4, and 5 respectively. (g) The impact of training memory without EM strategy. (h) The impact of training memory with EM strategy.}
 \label{fig:all_fig}
\end{figure*}

\begin{table*}[th]

    \centering
    \resizebox{0.83\textwidth}{!}{
    \tabcolsep=0.45cm 
    \renewcommand{\arraystretch}{0.85}
    \begin{tabular}{@{\extracolsep{4pt}}c|c|c|lcccccc@{}}
    \toprule
    \multirow{2}{*}{{Backbone}} & \multirow{2}{*}{{Method Type}} & Memory & \multirow{2}{*}{{Method}} & \multicolumn{2}{c}{\textbf{CIFAR10}} & \multicolumn{2}{c}{\textbf{CIFAR100}}  & \multicolumn{2}{c}{\textbf{ImageNet-R}} \\ \cline{5-6}  \cline{7-8}  \cline{9-10} \rule{0pt}{8pt}
      &  & Buffer &  & {Class-IL} & {Task-IL} &  {Class-IL} & {Task-IL} & {Class-IL} & {Task-IL} \\ \midrule

      \multirow{19}{*}{\Large{ResNet18}} & \multirow{2}{*}{-} & \multirow{2}{*}{-} & JOINT & 92.20 & 98.31 & 70.62 & 86.19  & 7.72 & 25.48 \\
      &  &  & FT-seq & 19.62 & 61.02 & 17.58 & 40.46 & 0.59 & 10.82 \\
      \cline{2-10} \rule{0pt}{8pt}
      & \multirow{3}{*}{\large{Non-Rehearsal based}} & \multirow{3}{*}{0} & EWC\cite{kirkpatrick2017overcoming} & 17.82 & 83.52 & 7.62 & 55.14 & 1.08 & 21.34 \\
      &  &  & LwF\cite{li2017learning} & 18.52 & 84.72 & 8.88 & 61.32 & 1.24 & 45.68 \\
      &  &  & SI\cite{zenke2017continual} & 18.41 & 84.74 & 6.73 & 50.44 & 3.31 & 22.72 \\
      \cline{2-10} \rule{0pt}{8pt}
      & \multirow{14}{*}{\large{Rehearsal based}} & \multirow{7}{*}{200} & ER\cite{riemer2018learning} & 44.79 & 91.19 & 21.40 & 61.36 & 1.01 & 15.36 \\
      &  &  & A-GEM \cite{chaudhry2018efficient} & 18.58 & 80.19 & 7.97 & 55.20 & 1.23 & 16.24 \\
      &  &  & iCaRL \cite{rebuffi2017icarl} & 23.80 & 67.82 & 7.31 & 33.10 & 0.81 & 9.20 \\
      &  &  & CVT \cite{ECCV22_CL} & 30.74 & 75.92 & 12.09 & 43.14 & 1.60 & 9.01 \\
      &  &  & SCoMMER \cite{sarfraz2023sparse} & 66.35 & 92.66 & 38.89 & 67.62 & 1.73 & 10.65 \\
      &  &  & DualNet \cite{pham2021dualnet} & 24.50 & 90.70 & 25.30 & 54.60 & 7.01 & 20.70 \\
      &  &  & BiMeCo \cite{nie2023bilateral} & 27.92 & 92.75 & 28.71 & 56.65 & 10.41 & 22.75 \\
      \cline{3-10} \rule{0pt}{8pt}
      &  & \multirow{7}{*}{500/600} & ER\cite{riemer2018learning} & 57.74 & 93.61 & 28.02 & 68.23 & 1.27 & 22.84 \\
      &  &  & A-GEM \cite{chaudhry2018efficient} & 24.85 & 84.80 & 8.89 & 51.47 & 1.23 & 19.35 \\
      &  &  & iCaRL \cite{rebuffi2017icarl} & 29.21 & 67.72 & 4.40 & 23.41 & 1.01 & 7.60 \\
      &  &  & CVT \cite{ECCV22_CL} & 40.13 & 79.61 & 13.83 & 46.39 & 1.24 & 6.97 \\
      &  &  & SCoMMER \cite{sarfraz2023sparse} & 73.95 & 94.14 & 49.09 & 74.50 & 1.40 & 10.05 \\
      &  &  & DualNet \cite{pham2021dualnet} & 35.00 & 91.90 & 34.65 & 62.70 & 8.70 & 20.40 \\
      &  &  & BiMeCo \cite{nie2023bilateral} & 38.40 & 93.95 & 38.05 & 64.75 & 12.13 & 22.45 \\ \hline

      \multirow{12}{*}{\Large{ViT}} & \multirow{2}{*}{-} & \multirow{2}{*}{-} & JOINT & 97.49 & 99.54 & 87.23 & 97.64 & 74.75 & 83.39 \\
      &  &  & FT-seq & 22.32 & 86.33 & 20.48 & 83.97 & 32.56 & 49.62 \\
      \cline{2-10} \rule{0pt}{8pt}
      & \multirow{2}{*}{\large{Non-Rehearsal based}} & \multirow{2}{*}{0} & L2P \cite{wang2022learning} & 92.22 & 98.99 & 79.68 & 96.24 & 48.68 & 65.38\\
      &  &  & DualPrompt \cite{wang2022dualprompt} & 94.43 & 99.32 & 79.98 & 95.92 & 52.20 & 69.22 \\
      \cline{2-10} \rule{0pt}{8pt}
      & \multirow{12}{*}{\large{Rehearsal based}} & \multirow{6}{*}{200} & L2P \cite{wang2022learning} & 67.13 & 96.39 & 65.29 & 92.16 & 36.70 & 55.35\\
      &  &  & DualPrompt \cite{wang2022dualprompt} & 70.60 & 97.78 & 65.97 & 92.85 & 38.79 & 59.32 \\
      &  &  & \cellcolor{gray!20} ICL w/o System2  & \cellcolor{gray!20} 94.60 & \cellcolor{gray!20} 99.43 & \cellcolor{gray!20} 77.34 & \cellcolor{gray!20} 94.81 & \cellcolor{gray!20} 49.87 & \cellcolor{gray!20} 68.62 \\
      &  &  & \cellcolor{gray!20} ICL w \textbf{MiniGPT4} & \cellcolor{gray!20} 95.34 & \cellcolor{gray!20} \underline{99.56} & \cellcolor{gray!20} 78.28 & \cellcolor{gray!20} 95.70 & \cellcolor{gray!20} 52.46 & \cellcolor{gray!20} 69.87 \\
      & & & \cellcolor{gray!20} ICL w \textbf{Inf-MLLM} & \cellcolor{gray!20} \underline{95.42} & \cellcolor{gray!20}	\underline{99.56} & \cellcolor{gray!20}	\underline{78.55} & \cellcolor{gray!20}	\underline{95.83} & \cellcolor{gray!20}	\underline{53.20}  & \cellcolor{gray!20}	\underline{72.96} \\
      & & & \cellcolor{gray!20} ICL w \textbf{Pure-MM} & \cellcolor{gray!20} \textbf{95.94} & \cellcolor{gray!20} 	\textbf{99.57} & \cellcolor{gray!20} 	\textbf{79.12}  & 	\cellcolor{gray!20}\textbf{95.99}  &  \cellcolor{gray!20}	\textbf{53.64}  & \cellcolor{gray!20}	\textbf{73.59} \\
      \cline{3-10} \rule{0pt}{8pt}
      &  & \multirow{6}{*}{500/600} & L2P \cite{wang2022learning} & 71.23 & 96.78 & 69.43 & 93.92 & 40.17 & 57.89\\
      &  &  & DualPrompt \cite{wang2022dualprompt} & 73.56 & 98.12 & 69.98 & 93.76 & 43.77 & 61.24 \\
       &  &  & \cellcolor{gray!20} ICL w/o System2 & \cellcolor{gray!20} 95.54 & \cellcolor{gray!20} 99.52 & \cellcolor{gray!20} 80.67 & \cellcolor{gray!20} 95.24 & \cellcolor{gray!20} 54.65 & \cellcolor{gray!20} 76.02 \\
       &  &  & \cellcolor{gray!20} ICL w \textbf{MiniGPT4} & \cellcolor{gray!20} 96.49 & \cellcolor{gray!20} 99.58 & \cellcolor{gray!20} 81.38 & \cellcolor{gray!20} 95.62 & \cellcolor{gray!20} 55.99 & \cellcolor{gray!20} 79.68 \\
        & & & \cellcolor{gray!20} ICL w \textbf{Inf-MLLM} & \cellcolor{gray!20} \underline{96.69}  & \cellcolor{gray!20}	\underline{99.64}  & \cellcolor{gray!20}	\underline{82.29}  & \cellcolor{gray!20}	\underline{96.14}  &  \cellcolor{gray!20}	\underline{57.47}  & \cellcolor{gray!20}	\underline{81.82} \\
        & & & \cellcolor{gray!20} ICL w \textbf{Pure-MM} & \cellcolor{gray!20} \textbf{96.83} & \cellcolor{gray!20}	\textbf{99.68}  & \cellcolor{gray!20}	\textbf{82.43}  & \cellcolor{gray!20}	\textbf{96.35} & \cellcolor{gray!20}	\textbf{58.18} & \cellcolor{gray!20}	\textbf{82.64} \\

      \bottomrule
    \end{tabular}}
    \caption{Comparison of average incremental accuracy (\%)
    with different continual learning method in two scenarios. Memory Buffer denotes the size of the memory buffer area ($0$ means no rehearsal is used). Note that when training with the ViT in System1, the buffer size of the Imagenet-R is $600$, due to the need to allocate an equal amount of buffer area for each class.}
    \label{tab1}
    \vspace{-16pt}
    \end{table*}

\subsection{Experimental Setups}
\vspace{-3pt}
Following \cite{van2019three, CVPR22_LVT}, we investigate two common CL setups:
Task-Incremental Learning (Task IL) and Class-Incremental Learning (Class IL). In Task IL, task identifiers are provided during both the training and testing phases. In contrast, the Class IL protocol assigns task identifiers only during the training phase. During the testing phase, the model faces the challenge of predicting all classes encountered up to that point, making it a more demanding scenario.

\noindent \textbf{Datasets.} 
Three datasets are used in our experiments:
CIFAR10 \cite{krizhevsky2009learning}, CIFAR100 \cite{krizhevsky2009learning}, ImageNet-R \cite{hendrycks2021many}. Details of these datasets are in Appendix.

\noindent \textbf{Baselines.} We combine the proposed ICL framework with the following advanced baselines:  Selected rehearsal-based methods: ER \cite{riemer2018learning}, A-GEM \cite{chaudhry2018efficient}, iCaRL \cite{rebuffi2017icarl}, CVT \cite{ECCV22_CL}, SCoMMER \cite{sarfraz2023sparse}, BiMeCo \cite{nie2023bilateral}. architecture-based methods: DualNet \cite{pham2021dualnet}, L2P \cite{wang2022learning}, DualPrompt \cite{wang2022dualprompt}. Additionally, we perform comparisons with well-known regularization-based techniques, namely EWC \cite{kirkpatrick2017overcoming}, LwF \cite{li2017learning}, and SI \cite{zenke2017continual}.
Additionally, we assess JOINT, which entails supervised fine-tuning across all task training sets and represents an upper performance limit. We also examine FT-seq, a sequential fine-tuning technique that partially freezes pre-training weights and generally serves as a performance lower bound. Both JOINT and FT-seq have two variants, one using ViT \cite{dosovitskiy2020image} and the other using ResNet18 \cite{he2016deep} as the backbone. Our main focus on comparing these methods with rehearsal- and prompt-based methods.

\noindent \textbf{Metrics.} We evaluate CL methods in terms of accuracy, following the accuracy definition presented in \cite{CVPR22_LVT, wang2022dualprompt, buzzega2020dark}: $\mathbf{A}_{T}=\frac{1}{T}\sum_{t=1}^{T}a_{T,t}$, where $a_{T, t}$ represents the testing accuracy for task $\mathcal{T}_t$ when the model has completed learning task $\mathcal{T}_T$.

\noindent \textbf{Implementation Details.} Following \cite{ECCV22_CL, rebuffi2017icarl, buzzega2020dark, buzzega2021rethinking, zenke2017continual}, we use RestNet18 \cite{he2016deep}, ViT \cite{dosovitskiy2020image} as the backbone in System1. For System2, we choose MiniGPT4 \cite{zhu2023minigpt}, Inf-MLLM \cite{zhou2023infmllm}, Pure-MM as the backbone. 
More details are in Appendix.

\vspace{-5pt}
\subsection{Results}
\vspace{-3pt}
Extensive experiments are conducted on CIFAR10, CIFAR100, ImageNet-R. Specifically, we add our ICL to several state of the arts methods to evaluate its effectiveness. 

\noindent  \textbf{Results of comparisons with State-of-the-Art Methods. }
The quantitative comparisons are summarized in Tab. \ref{tab1}.
In the rehearsal-based method, we experiment with different buffer sizes for comparison. The results consistently show that our method outperforms others. Notably, to better simulate the CL scenario, we restrict the number of epochs to one, allowing the model to encounter the data only once during incremental task learning. This restriction significantly affects the efficacy of regularization techniques like EWC \cite{kirkpatrick2017overcoming}, LwF \cite{li2017learning}, as well as rehearsal-based methods such as ER \cite{riemer2018learning}, A-GEM \cite{chaudhry2018efficient}, iCaRL \cite{rebuffi2017icarl} and CVT \cite{ECCV22_CL}, among others. However, transformer-based approaches like L2P \cite{wang2022learning} and DualPrompt \cite{wang2022dualprompt} maintain certain performance.
Additionally, even in the absence of integrating the System2, its incorporation leads to further performance enhancement. This is demonstrated by a notable increase of over 3\% in CL accuracy on the ImageNet-R dataset, observed across 10 consecutive splits, with a memory capacity of 600. The introduction of the System2 enhances System1 ability to effectively recognize and address previously forgotten images or information.
To ensure fair comparisons, we incorporate a buffer into the L2P \cite{wang2022learning} and DualPrompt \cite{wang2022dualprompt} methods, using a fixed strategy akin to CVT \cite{ECCV22_CL}. Subsequently, we observe that the performance of the L2P \cite{wang2022learning} and DualPrompt \cite{wang2022dualprompt} methods declined after integrating the buffer. This decline is likely due to interference from the replayed samples in the buffer, which hampers the training of task-specific prompts and leads to greater forgetting.
\vspace{-2pt}
\noindent\textbf{Results of different task settings. } To assess CL strategies across varying numbers of data streams, following the protocol outlined in \cite{ECCV22_CL, yan2021dynamically}. This method entails dividing the ImageNet-R dataset, which consists of $200$ classes, into subsets containing $5$, $10$, and $20$ classes each, thereby creating incremental tasks.
Tab. \ref{tab2} offers a comprehensive analysis of the accuracy achieved by various methods across different task configurations. The results unequivocally establish the significant superiority of our method over both regularization-based and rehearsal-based method in a wide range of incremental division scenarios. Notably, even when the memory capacity is held constant at $200$, our method outperforms architecture-based methods such as L2P \cite{wang2022learning} and DualPrompt \cite{wang2022dualprompt}. Furthermore, the results illustrate a consistent enhancement in our method performance as the buffer size increases.

\begin{table}[t]
\centering
\resizebox{0.423\textwidth}{!}{%
\renewcommand\arraystretch{0.85}
\tabcolsep=0.01cm
\begin{tabular}{@{\extracolsep{4pt}}lccccccc@{}}
\toprule
Memory &\multirow{2}{*}{{Method}} & \multicolumn{2}{c}{\textbf{5 splits}} & \multicolumn{2}{c}{\textbf{10 splits}} & \multicolumn{2}{c}{\textbf{20 splits}} \\ \cmidrule{3-4} \cmidrule{5-6} \cmidrule{7-8}
Buffer &  &  {Class-IL} & {Task-IL} &  {Class-IL} & {Task-IL} & {Class-IL} & {Task-IL} \\ \midrule
 
 \multirow{5}{*}{0} & EWC\cite{kirkpatrick2017overcoming} & 1.56 & 11.35 & 1.08 & 21.34 & 8.10 & 12.68 \\
 & LwF\cite{li2017learning} & 1.38 & 14.66 & 1.24 & 45.68 & 0.77 & 14.34 \\
 & SI\cite{zenke2017continual} & 1.78 & 11.50 & 3.31 & 22.72 & 3.35 & 40.29 \\
 & L2P \cite{wang2022learning} & 29.87 & 38.58 & 48.68 & 65.38 & 20.08 & 47.98 \\
 & DualPrompt \cite{wang2022dualprompt} & 54.43 & 66.86 & 52.20 & 69.22 & 47.13 & 71.43 \\ \midrule
 
 \multirow{9}{*}{200} & ER\cite{riemer2018learning} & 1.29 & 9.75 & 1.01 & 15.36 & 1.38 & 19.81 \\ 
 & A-GEM \cite{chaudhry2018efficient} & 1.30 & 4.23 & 1.23 & 16.24 & 1.34 & 21.78 \\ 
 & iCaRL \cite{rebuffi2017icarl} & 0.41 & 3.22 & 0.81 & 9.20 & 0.83 & 15.23 \\ 
 & CVT \cite{ECCV22_CL} & 1.47 & 5.65 & 1.6 & 9.01 & 1.01 & 13.19 \\
 & SCoMMER \cite{sarfraz2023sparse} & 0.80 & 3.78 & 1.73 & 10.65 & 0.32 & 10.36 \\ 
 & DualNet \cite{pham2021dualnet} & 9.32 & 13.14 & 7.01 & 20.70 & 4.92 & 25.53 \\ 
 & BiMeCo \cite{nie2023bilateral} & 11.18 & 14.27 & 10.41 & 22.75 & 5.86 & 26.33 \\ 
 & \cellcolor{gray!20} ICL w/o System2 (ours) & \cellcolor{gray!20} 49.91 & \cellcolor{gray!20} 74.36 & \cellcolor{gray!20} 49.87 & \cellcolor{gray!20} 68.62 & \cellcolor{gray!20} 48.75 & \cellcolor{gray!20} 73.95\\ 
 & \cellcolor{gray!20} ICL (ours) & \cellcolor{gray!20} 54.85 & \cellcolor{gray!20} \underline{75.57} & \cellcolor{gray!20} 52.46 & \cellcolor{gray!20} 69.87 & \cellcolor{gray!20} 49.98 & \cellcolor{gray!20} 74.63\\
 \midrule

 \multirow{9}{*}{500/600} & ER\cite{riemer2018learning} & 1.30 & 13.02 & 1.27 & 22.84 & 1.56 & 20.42 \\ 
 & A-GEM \cite{chaudhry2018efficient} & 1.31 & 9.55 & 1.23 & 19.35 & 1.58 & 21.89 \\ 
 & iCaRL \cite{rebuffi2017icarl} & 0.43 & 3.62 & 1.01 & 7.60 & 1.62 & 14.23 \\ 
 & CVT \cite{ECCV22_CL} & 1.95 & 5.68 & 1.24 & 6.97 & 1.45 & 15.58 \\
 & SCoMMER \cite{sarfraz2023sparse} & 0.61 & 3.35 & 1.40 & 10.05 & 0.56 & 12.60 \\
 & DualNet \cite{pham2021dualnet} & 10.03 & 13.44 & 8.70 & 20.40 & 6.40 & 31.80 \\
 & BiMeCo \cite{nie2023bilateral} & 11.89 & 14.57 & 12.13 & 22.45 & 7.34 & 32.73 \\
 & \cellcolor{gray!20} ICL w/o System2 (ours) & \cellcolor{gray!20} \underline{54.60} & \cellcolor{gray!20} \underline{75.57} & \cellcolor{gray!20} \underline{54.65} & \cellcolor{gray!20} \underline{76.02} & \cellcolor{gray!20} \underline{52.46} & \cellcolor{gray!20} \underline{77.05}\\ 
 & \cellcolor{gray!20} ICL (ours) & \cellcolor{gray!20} \textbf{56.34} & \cellcolor{gray!20} \textbf{78.36} & \cellcolor{gray!20} \textbf{55.99} & \cellcolor{gray!20} \textbf{79.68} & \cellcolor{gray!20} \textbf{53.60} & \cellcolor{gray!20} \textbf{80.47}\\
 
\bottomrule
\end{tabular}}
\caption{Comparison under different task number settings on ImageNet-R dataset.}
\vspace{-16pt}
\label{tab2}
\end{table}

\noindent \textbf{Results of forgetting curve comparison.} 
To illustrate the forgetting process within each compared methods in the CL data stream, we record the average test accuracy for both the current and preceding tasks upon completing the training of each task. Subsequently, we create a line chart to visualize the change in accuracy with the addition of each task, offering a visual representation of the forgetting process.
Fig. \ref{fig:all_fig}(a) and Fig. \ref{fig:all_fig}(b) provide clear illustrations that, as new tasks are introduced, most methods exhibit a decline in performance. However, our method consistently outperforms these methods in terms of accuracy at every stage.

\vspace{-5pt}
\subsection{Ablation Study}
\vspace{-3pt}

\noindent \textbf{Analysis of hyperparameters $\lambda$ and margin $\delta$. }

Fig. \ref{fig:all_fig}(c) and Fig. \ref{fig:all_fig}(d) depict the impact of parameters $\lambda$ and $\delta$ on our method performance, using the CIFAR100 and ImageNet-R datasets, with memory sizes set at 500 and 600, respectively. An analysis of the heat maps reveals that our method's sensitivity to $\lambda$ varies between datasets. Specifically, CIFAR100 demonstrates optimal performance when $\lambda=0.1$, while ImageNet-R maintains consistent performance with $\lambda$ values within the range $[0.001, 0.01, 0.1]$. In contrast, variations in $\delta$ do not result in substantial accuracy differences between the datasets. This suggests that the margin regularization term has only a marginal impact on overall model performance. Importantly, when $\lambda = 0$ and $\delta = 1$ (as shown in the bottom), the absence of the regularization term leads to diminished performance, underscoring the value of incorporating margin regularization to enhance the model's effectiveness.

\noindent \textbf{Analysis of concentration parameter Selection. } We conduct an investigation into the influence of the concentration parameter $\kappa$ on model performance. Fig. \ref{fig:all_fig}(e) indicates that there is no significant variation in performance across different values of $\kappa$ for the CIFAR10 dataset. However, for datasets such as CIFAR100 and ImageNet-R, it becomes important to estimate the concentration in advance to ensure a more accurate modeling of the concentration. Consequently, these datasets demonstrate a noticeable sensitivity to the concentration parameter.

\noindent \textbf{Impact of Top-k results in collaboration inference.}
To investigate the influence of the number of prompt categories $K$ on the reasoning process of System2, a series of experiments are conducted using different values of $K$ on multiple datasets, as depicted in Fig.
\ref{fig:all_fig}(f). The results clearly indicate that memory vectors possessing well-defined geometric structures facilitate stable memory retrieval, thereby preventing System1 from deviating significantly from the central data point during the reasoning phase. Consequently, when a smaller number of top-k choices are provided, it can ensure almost guaranteed inclusion of the correct categories. Conversely, when $K$ is larger, there is an increased tendency for erroneous category information to be included in the prompt, further intensifying the potential for confusion in System2. Fig. \ref{fig:all_fig}(f) also presents the accuracy achieved without System2 across various datasets (blue horizontal line in the figure), demonstrating the performance enhancement by the incorporation of System2.

\noindent \textbf{Impact of separate query and value memory optimization. }
To assess the query-value separation strategy impact on stable value memory modeling in persistent scenarios, we compare its use (Fig. \ref{fig:all_fig}(g)) and absence (Fig. \ref{fig:all_fig}(h)). Visualizing the value memory via tsne reduction reveals that query-value parameter optimization can create more focused value memory modeling, enhancing task discrimination. The corresponding results consistently validate the effectiveness of our proposed CL-vMF mechanism.

\vspace{-5pt}
\section{Conclusion}
\vspace{-5pt}

In our paper, we introduced ICL, a groundbreaking continual learning (CL) paradigm inspired by Complementary Learning System theory in neurocognitive science. ICL combines ViT with an interactive query and value memory module powered by CKT-MHA, enhancing the efficiency of fast thinking (System1). Additionally, it leverages our CL-vMF mechanism to improve memory representation distinction. ICL also integrates multi-modal Large Language Models (System2) with System1 for advanced reasoning, dynamically modulated by hard examples detected through our VMF-ODI strategy.
Our experiments confirmed the effectiveness of our framework in reducing forgetting, surpassing contemporary state-of-the-art methods.  

\textbf{Acknowledgement} This work was supported in part by the National Key R\&D Program of China (No. 2023YFC3305102). We extend our gratitude to the anonymous reviewers for their insightful feedback, which has greatly contributed to the improvement of this paper.
{
    \small
    \bibliographystyle{ieeenat_fullname}
    \bibliography{main}
}
\clearpage
\setcounter{page}{1}
\maketitlesupplementary

\section{Algorithms of ICL}
We formalize the algorithms for the training and inference stages of ICL, as shown in Algorithm \ref{Algo1} and \ref{Algo2}. Here, we set the detection threshold $\alpha$ as the upper $20$th percentile of the standard normal distribution, which is $-0.842$.
\begin{algorithm}
\caption{Training Stage of ICL}          
\begin{algorithmic}[1]
\REQUIRE The parameters of the image feature extractor of the pre-trained ViT $\boldsymbol\varphi$, the memory buffer $\mathcal M$ with size $
|\mathcal M|$, the parameters of the query memory $\boldsymbol\theta$, the CL training dataset $\mathcal D_t, 1 \leq t \leq \mathcal |T|$ with $n^t$ batches
\STATE{Initialize value memory parameters $\mathcal Z=\emptyset$ and memory buffer $\mathcal M=\emptyset$.}
\FOR{$t=1\leftarrow |T|$}
\FOR{$i=1\leftarrow n^t$}
\IF{$t>1$}
\STATE Randomly sample a batch $\tilde{\mathcal B}^t_i$ from $\mathcal M$
\STATE $\mathcal B^t_i = \mathcal B^t_i\cup \tilde{\mathcal B}^t_i$
\ENDIF
\IF{$\exists\left(\boldsymbol x^t, y^t\right)\in\mathcal B^t_i$ s.t. no $\boldsymbol z\in\mathcal Z$ matches $y^t$}

\STATE Add $\boldsymbol z^{y^t} = \text{Concat}[\boldsymbol z_t, \boldsymbol{z}_{y^t}]$ into $\mathcal Z$.
\ENDIF
\STATE E-step: Update value memory parameters $\mathcal Z$ on $\mathcal L(\mathcal B^t_i)$
\STATE M-step: Update query memory parameters $\boldsymbol \theta$ on $\mathcal L(\mathcal B^t_i)$
\STATE Update memory buffer $\mathcal M$
\ENDFOR
\STATE Freeze the memory parameters of classes in task $t$
\ENDFOR
\end{algorithmic}
\label{Algo1} 
\end{algorithm}

\begin{algorithm}
\caption{Inference Stage of ICL}          
\begin{algorithmic}[1]
\REQUIRE The image feature extractor in pre-trained ViT $f_{\boldsymbol\varphi}$, the trained query and value memory $f_{\boldsymbol\theta}$, $\mathcal Z$, the test dataset $\mathcal D$ with $n$ batches.

\FOR{$i=1\leftarrow n$}
\FOR{$j\leftarrow|\mathcal B_i|$} 
\STATE{$\hat y_j = \arg\max_{y}p^{\mathcal B^t_i}_{\boldsymbol{\theta},\boldsymbol \varphi}(\boldsymbol z^{y_i}|\boldsymbol x)$}
\ENDFOR
\STATE{$\tilde {\boldsymbol{X}_i} = \left\{(\tilde{\boldsymbol x},\tilde y )\in\mathcal B_i\mid (\nu-\bar\nu_{\mathcal B_i})/\sigma_{\mathcal B_i}<\alpha\right\}$}
\IF{$\tilde {\boldsymbol{X}_i} = \emptyset$}
\STATE{Return the prediction results of System 1}
\ELSE
\STATE{Using System 2, perform inference on $\boldsymbol x\in \tilde {\boldsymbol{X}_i}$ by combining the top-$K$ output of System 1. Retrieve the result of the exact answer and combine it with the remaining predictions from System 1 before returning.}
\ENDIF
\ENDFOR
\end{algorithmic}
\label{Algo2} 
\end{algorithm}

\section{Datasets Settings}

CIFAR-10 comprises 10 classes, each with 50,000 training and 10,000 test color images. CIFAR-100 includes 100 classes, offering 500 training and 100 testing images per class. ImageNet-R, an extension of the ImageNet dataset, possess 200 classes. It contains a total of 30,000 images, of which $20\%$ were allocated as the test set.

CIFAR-10 was divided into five tasks, two classes allocated to each task. CIFAR-100 was divided into ten tasks, each task with ten classes. Similarly, ImageNet-R was organized into ten tasks, with each task containing 20 classes. Input images were resized to $224 \times 224$ and normalized to the range $[0,1]$. ICL was compared against both representative baselines and state-of-the-art methods across diverse buffer sizes and datasets.

\section{Implementation Details}

To ensure a fair comparison between methods, we carried out uniform resizing of the images to dimensions of $224 \times 224$ and applied image normalization. Following the settings of \cite{ECCV22_CL, rebuffi2017icarl, buzzega2020dark, buzzega2021rethinking}, we adopted 10 batch size and 1 epoch for all methods during training, utilizing cross-entropy as the classification loss. For the L2P\cite{wang2022learning}, DualPrompt\cite{wang2022dualprompt} approaches, we followed the implementation details of the original paper and employed ViT as the backbone network, while ResNet18 served as the backbone network for the remaining methods. We meticulously reproduced the outcomes by adhering to the original implementation and settings. We have set up separate Adam optimizers with a constant learning rate of $1e-4$ for the query and value memory parameters.

\section{Inference with System 1}
We conducted a comparison by directly applying rehearsal-based fine-tuning with a same-sized buffer, using only the pretrained ViT with a trainable classification head on each dataset. The results, as shown in Tab. \ref{withvit}, were significantly lower than those obtained using only System 1. This stark contrast serves as strong evidence that both ViT and MiniGPT-4 have not undergone pretraining on the three datasets and highlights the effectiveness of our proposed method.

\begin{table}[h]
    \centering
    \renewcommand\arraystretch{0.55}
    \resizebox{0.47\textwidth}{!}{%
    \begin{tabular}{cccccccc}
    \toprule
    Memory & \multirow{2}{*}{{Method}} & \multicolumn{2}{c}{\textbf{CIFAR10}} & \multicolumn{2}{c}{\textbf{CIFAR100}}  & \multicolumn{2}{c}{\textbf{ImageNet-R}} \\ \cline{3-4}  \cline{5-6}  \cline{7-8} \rule{0pt}{8pt}
    Buffer &  & {Class-IL} & {Task-IL} &  {Class-IL} & {Task-IL} & {Class-IL} & {Task-IL} \\ \midrule
    \multirow{4}{*}{{200}} & ViT Finetune &  33.15 & 	96.00 & 	32.60 & 	91.50 & 	20.88 & 	64.45 \\
     & ICL w/o System2 & 94.60 & 	99.43 & 	77.34 & 	94.81 & 	49.87 & 	68.62 \\ \midrule
     \multirow{4}{*}{{500/600}} & ViT Finetune &  62.65 & 	97.15 & 	45.30  & 	92.80 & 	33.26 & 	75.50  \\
     & ICL w/o System2 & 95.54 & 	99.52 & 	80.67 & 	95.24 & 	54.65 & 	76.02  \\
    
    \bottomrule
    \end{tabular}}
    \vspace{-3pt}
    \caption{Comparison of incremental accuracy (\%). Vit Finetune represents the basic rehearsal method using Vit as the backbone.}
    \vspace{-10pt}
    \label{withvit}
\end{table}

\section{Inference with System 2}

In order to validate the mutually beneficial interaction between systems, we conduct experiments using the pre-trained MiniGPT4 \cite{zhu2023minigpt} to perform inference on the test sets of CIFAR-10, CIFAR-100, and ImageNet-R. MiniGPT4 loads the official 7B pre-trained parameters, and the prompt used by MiniGPT4 is the same as System2. Since System 1 does not provide a topk option, we provided all categories to MiniGPT4, allowing it to select a category for image classification based on the image description. Tab. \ref{supp exp} presents the accuracy of reasoning, error rate, and proportion of no exact response (i.e. there is not only one class in the given classes is returned or no response).

\begin{table}[th]
    \centering
    \resizebox{0.47\textwidth}{!}{%
    \begin{tabular}{ccccc}
    \toprule
        Dataset & Accuracy & Error & No Response & Total \\
        \midrule
        CIFAR-10 & 9.53 & 15.04 & 75.43 & 10000 \\
        CIFAR-100 & 2.45 & 14.53 & 83.02 & 10000 \\
        ImageNet-R & 2.67 & 10.33 & 87.00 & 6000 \\
        \bottomrule
    \end{tabular}}
    \caption{MiniGPT4's inference accuracy, error rate, and proportion of no exact response on the CIFAR-10, CIFAR-100, and ImageNet-R test sets. The number of responses for each test set are reported in the last column. }
    \label{supp exp}
\end{table}

The results presented in the table indicate that over $75\%$ of the images fed in MiniGPT4, when applied to the CIFAR-10 dataset, fail to return a specific class to which the image belongs. And when faced with the CIFAR-100 and ImageNet-R datasets, MiniGPT4 with prompt that includes a larger number of classes, encounters increased difficulty in making accurate selections. Among the images that were returned with specific class information, over two-thirds were misclassified. These experimental results demonstrate that relying solely on MiniGPT4 for image classification tasks yields poor performance. Nevertheless, when System 1 offers the top-K option, incorporating MiniGPT4 as System2 enhances the image classification task and improves the final accuracy. This finding demonstrates that the interactive inference between System1 and System2 enables mutual promotion and improvement.

The limitations of MiniGPT-4 restricted the performance enhancement of System 2. To address these concerns, we chose more advanced MLLMs as System 2. As depicted in Tab. \ref{tab1}, there was a notable 3-4\% improvement, especially on the challenging ImageNet-R dataset.

\end{document}